\documentclass[letterpaper, 10 pt, conference]{ieeeconf}  

\IEEEoverridecommandlockouts                              

\usepackage{amsmath} 
\usepackage{svg}
\usepackage{booktabs}
\usepackage{xspace}
\usepackage{color}
\usepackage{url}
\usepackage{afterpage}
\definecolor{myblue}{rgb}{0.239,0.553,0.565}

\newcommand{\cocreate}{I_{cpm}\xspace}
\newcommand{\frida}{I_{sim}\xspace}
\newcommand{\canvascontinuation}{co-painting\xspace}
\newcommand{\Canvascontinuation}{Co-Painting\xspace}

\title{\LARGE \bf
CoFRIDA:   Self-Supervised Fine-Tuning for Human-Robot \Canvascontinuation
}

\author{Peter Schaldenbrand$^{1}$, Gaurav Parmar$^{1}$, Jun-Yan Zhu$^{1}$, James McCann$^{1}$, and Jean Oh$^{1}$%
    \thanks{
        $^{1}$The Robotics Institute, Carnegie Mellon University
    }
    \thanks{
        \{pschalde, jmccann, gparmar, junyanz, hyaejino\}@andrew.cmu.edu
    }
}

\begin{document}

\maketitle
\thispagestyle{empty}
\pagestyle{empty}

\begin{abstract}

Prior robot painting and drawing work, such as FRIDA, has focused on decreasing the sim-to-real gap and expanding input modalities for users, but the interaction with these systems generally exists only in the input stages. 
To support interactive, human-robot collaborative painting, we introduce the Collaborative FRIDA (CoFRIDA) robot painting framework,
which can \textit{co-paint} by modifying and engaging with content already painted by a human collaborator. 
To improve text-image alignment--FRIDA's major weakness--our system uses pre-trained text-to-image models; however, pre-trained models in the context of real-world co-painting do not perform well because they (1) do not understand the constraints and abilities of the robot and (2) cannot perform co-painting without making unrealistic edits to the canvas and overwriting content.
We propose a self-supervised fine-tuning procedure that can tackle both issues, allowing the use of pre-trained state-of-the-art text-image alignment models with robots to enable co-painting in the physical world.
Our open-source approach, CoFRIDA, creates paintings and drawings that match the input text prompt more clearly than FRIDA, both from a blank canvas and one with human created work. More generally, our fine-tuning procedure successfully encodes the robot's constraints and abilities into a foundation model, showcasing promising results as an effective method for reducing sim-to-real gaps. 
\url{https://pschaldenbrand.github.io/cofrida/}

\end{abstract}

\section{INTRODUCTION}

\begin{figure}[ht!]
    \centering
    \includegraphics[width=\columnwidth]{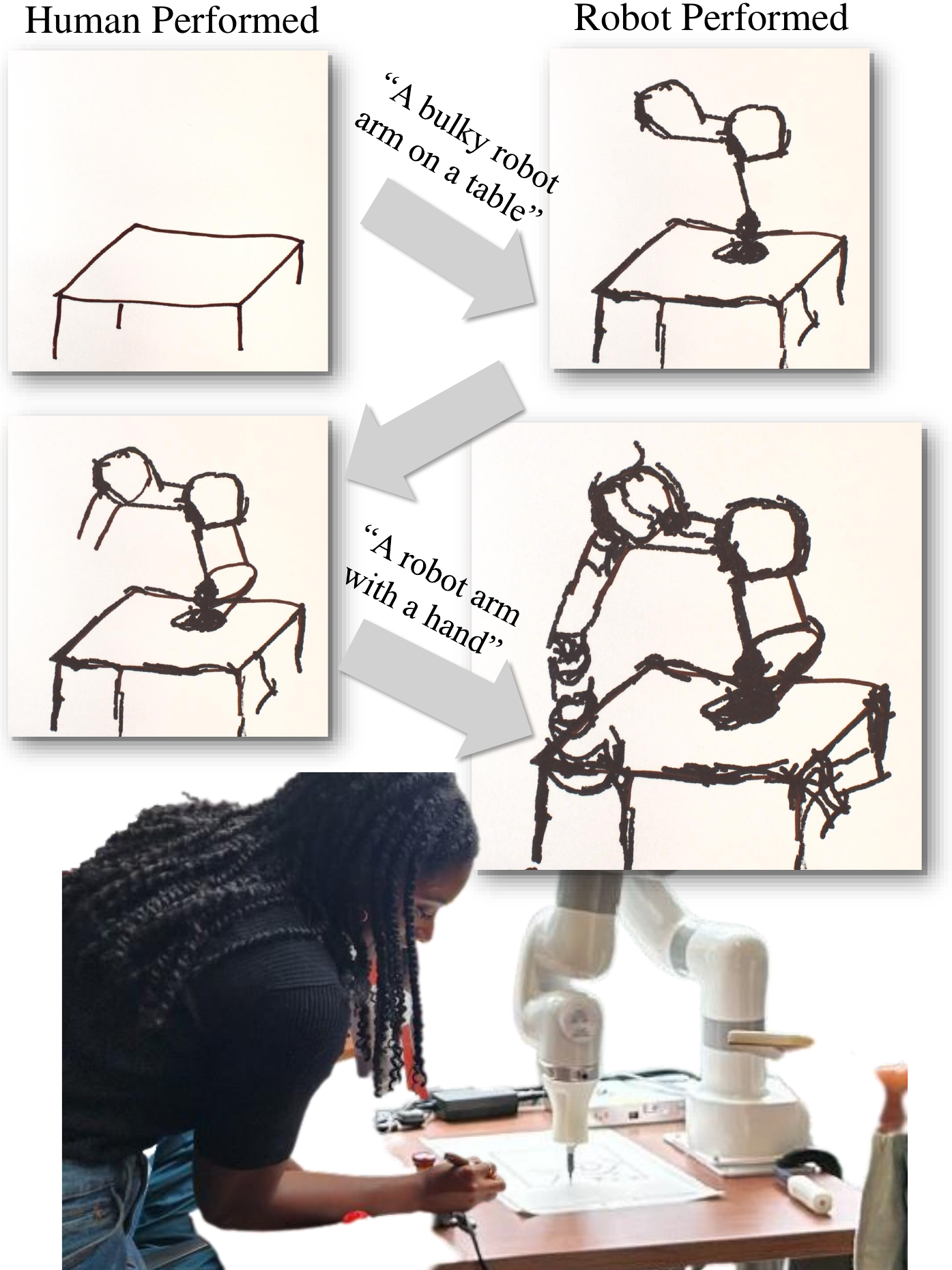}
    \vspace{-5mm}
    \caption{
    \textbf{Co-Painting with CoFRIDA.} 
    We showcase how CoFRIDA collaboratively paints with artists.
    The process begins with the artist sketching a table.
    Building on that foundation, CoFRIDA adds to the canvas, guided by the artist's initial prompt: ``A bulky robot arm on a table."
     The artist then iterates on the painting with additional strokes to add detail to the robot arm, and provides a new text prompt, ``A robot arm with a hand." CoFRIDA responds by completing the painting to match this new description.
    }
    \vspace{-5mm}
    \label{fig:main_fig}
\end{figure}

While recent breakthroughs in text-to-image synthesis technologies have ignited a boom in digital content generation, using them to produce art with robots is still in its infancy due to a significant gap between simulated and real-world environments.
FRIDA~\cite{schaldenbrand2023frida} is a robotic framework that can take user inputs, such as language descriptions or input images, to paint on a physical canvas using a paintbrush and acrylic paint. While FRIDA aims at giving users control over content generation, the users are allowed to add their input only with an initial input prompt or image, after which they are excluded from the creative process. While it is still debatable whether such an autonomous creation is desired by humans practicing art~\cite{gebru2023artAiImpact}, there is strong evidence of the potential value of a co-creative agent~\cite{bateman2021creating, davis2016DrawingApprentice, ibarrola2023collaborative, lawton2023reframer, lawton2023toolAtool,jansen2021coDrawWorkflows, lee2024adversarialRobot} specifically in the domain of art therapy~\cite{cooney2018artTherapyRobots, cooney2019artTherapyRobots, cooney2021robot, shaik2021coCreativeDiffAbledKids}. The benefits can be further increased when paired with a physical embodiment of such an agent and drawing in the real world~\cite{lin2020cobbie, herath2020artsHealth}. 
To invite users into the creative process and bring the benefits of both co-creation and robotic embodiment, we build on FRIDA to propose a Collaborative Framework and Robotics Initiative for Developing Arts (CoFRIDA), as illustrated in Fig.~\ref{fig:main_fig}.

One of the biggest challenges in human-robot co-creation is enabling the robot to create new content that engages with the existing content that the human drew, hereby referred to as \textit{\canvascontinuation}. 
While there exist related image editing problems, such as in-painting, \canvascontinuation is a new class of problems with unique challenges as it is undesirable in \canvascontinuation to make radical changes to the image that would overwrite the human's previous work.
In in-painting, the area for editing an image is coarsely specified by the user and the model is expected to drastically change the content within that local region. By contrast, with \canvascontinuation, the edit is expected to preserve and engage with the full canvas rather than re-imagining a local region. Whereas in-painting is a localized edit by definition, \canvascontinuation is a continuous, iterative completion, e.g., adding detail to an existing human-drawn rough sketch.

Besides the challenges of \canvascontinuation, robotic image creation is difficult due to real-world constraints, such as existing canvas state, limited abilities of the robot, tools and materials available to the robot, and stochasticity in robot performance.
These robotic constraints vastly limit the content that is capable of being created, as illustrated in the left side of Fig.~\ref{fig:semantic_sim2real_gap2}. With a large paintbrush, fine-details are not achievable, and with a single marker, multi-color images are not possible. Multiple works address these constraints to decrease the Sim2Real gap, but only paint from image inputs~\cite{schaldenbrand2021contentMaskedLoss, lindemeier2018Edavid, wang2020calligraphy, carter2017DarkFactoryPortraits}. 
Even fewer existing works use cameras to enable co-creation of images~\cite{lin2020cobbie}.

FRIDA uses the data from a real robot to be able to simulate high-fidelity brush strokes using the idea known as Real2Sim2Real. To paint from a language input, FRIDA uses CLIP~\cite{radford2021-clip} to align language and image which tends to generate noisy output. To improve the quality of paintings for CoFRIDA, we use powerful image generators pre-trained using gigantic text-image paired data, e.g., StableDiffusion~\cite{rombach2021stableDiffusion} or Instruct-Pix2Pix~\cite{brooks2023instructpix2pix}.   
Because such pre-trained image generators do not know the capabilities of the robot, there is both a large difference in pixel value and semantic meaning between the image generator output and FRIDA's simulated plan. The former difference is a traditional Sim2Real gap, whereas the latter is a concept we introduce as the \textit{Semantic Sim2Real Gap}.

To reduce the Semantic Sim2Real Gap, we propose a self-supervised fine-tuning for CoFRIDA. 
CoFRIDA adapts a pre-trained image generator to both generate content within the abilities of the robot and perform \canvascontinuation to enable human-robot collaborative drawing from language guidance, e.g., in this paper, we use Instruct-Pix2Pix~\cite{brooks2023instructpix2pix} as our base text-image model. 
To adapt a pre-trained model for \canvascontinuation and encode robotic constraints, first we create the self-supervised fine-tuning dataset by using FRIDA to generate full drawings or paintings of images from a text-image dataset. Strokes from the full paintings are removed selectively to form partial paintings. We fine-tune Instruct-Pix2Pix by retraining it with a low learning rate to predict the full painting from the partial painting and text prompt.

\begin{figure}[t!]
    \centering  \small
    \includegraphics[width=\columnwidth]{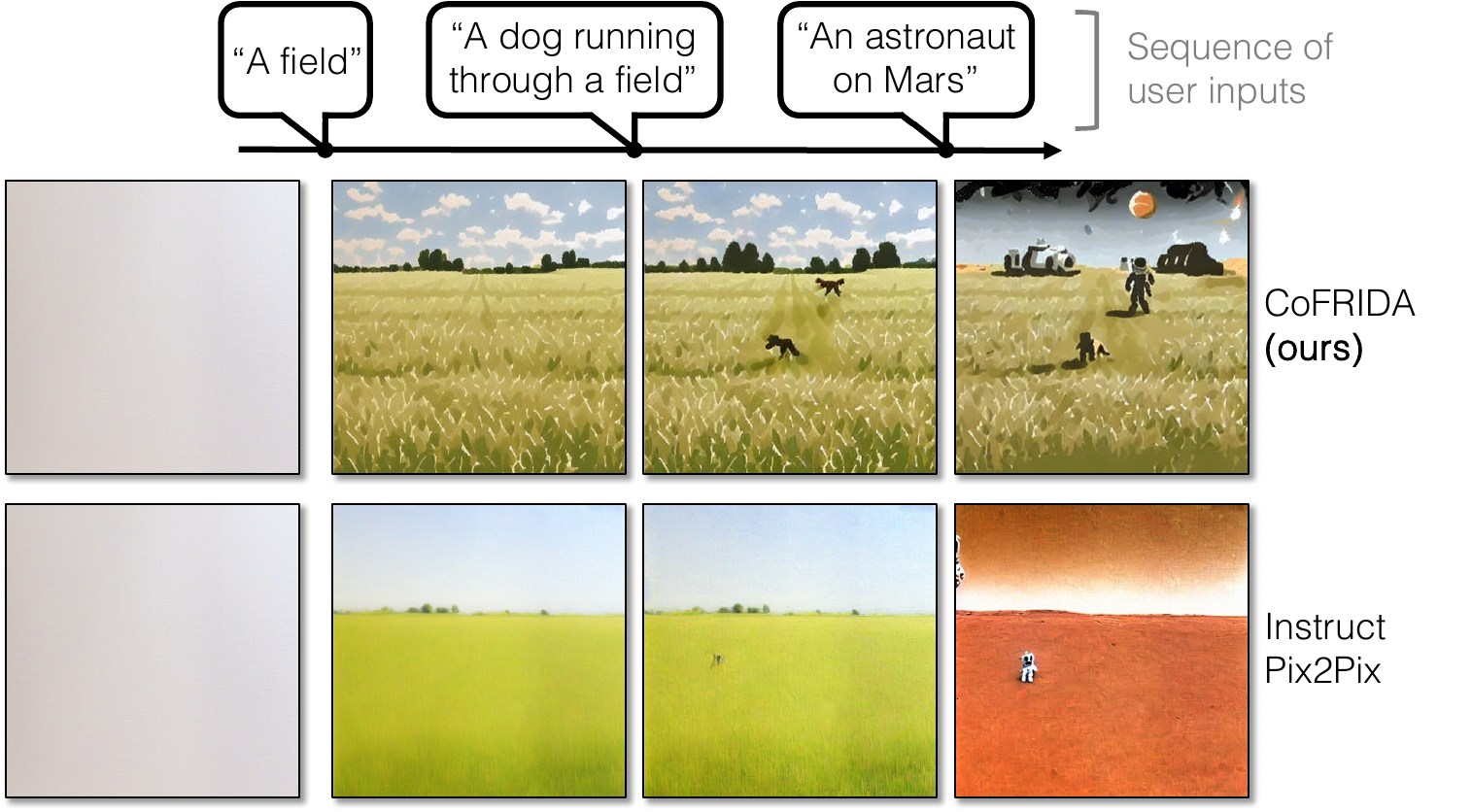}
    \vspace{-6mm}
    \caption{
    \textbf{\Canvascontinuation.} 
    We introduce Co-Painting as a task in which a robot must add content to a painting that engages with the current content without destroying the existing work. We demonstrate that existing models (Instruct-Pix2Pix, bottom row) often cannot successfully add content without making unreasonably large edits to the canvas, overwriting any prior work, while CoFRIDA (top row) adds content that harmonizes with the existing work.
    }
    \label{fig:multiple_runs}
    \vspace{-7mm}
\end{figure}


CoFRIDA can successfully use an existing canvas state to generate future actions towards a language goal without completely overwriting the existing work as shown in Fig.~\ref{fig:multiple_runs}. Based on a survey on Amazon Mechanical Turk (MTurk) of 24 participants, CoFRIDA's completed drawings from partial sketches were found to be substantially more similar to the language goal when compared to those by the baselines.

Our main contributions are summarized as: 1) we introduce \canvascontinuation, a new class of image editing that is required for human-robot collaborative creation; 2) we propose Collaborative FRIDA (CoFRIDA) to support human-robot co-painting to produce real-world arts, e.g., paintings and drawings on canvas; 3) we propose a generalizable method for reducing the Sim2Real gap using self-supervised fine-tuning, enabling generic pre-trained models to be used with physical robots; 4) CoFRIDA is open-source\footnote{\url{https://github.com/cmubig/Frida}} and available on XArm, Franka Emika, and Rethink Sawyer robot platforms.

\begin{figure*}[t!]
    \centering \small
    \includegraphics[width=\textwidth]{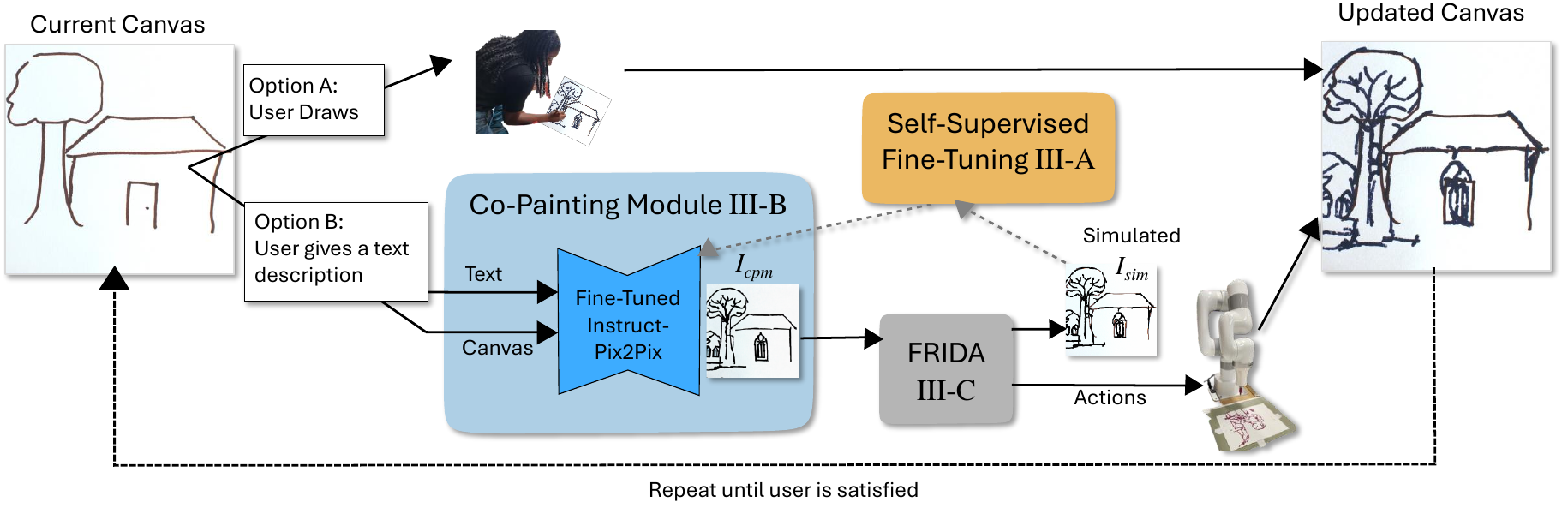}
    \vspace{-7mm}
    \caption{
    \textbf{Method Overview.}
    Offline, we fine-tune a pre-trained Instruct-Pix2Pix model on our self-supervised data.
    Online, the user can either draw or give the robot a text description. The Co-Painting Module takes as input the current canvas and text description to generate a pixel prediction of how the robot should finish the painting using the fine-tuned Instruct-Pix2Pix model. 
    FRIDA predicts actions for the robot to create this pixel image and produces a simulation. This process is repeated until the user is satisfied.}
    \label{fig:method}
    \vspace{-5mm}
\end{figure*}

\vspace{-9pt}
\section{RELATED WORK}

\subsection{Computer-Based Image Co-Creation}

Computer-based image co-creation generally involves turn taking between a human and a computer in applying brush stroke primitives towards one of a discrete set of goals, as in \texttt{sketch-rnn}~\cite{ha2017sketchRnn} and Drawing Apprentice~\cite{davis2016DrawingApprentice}, or even towards natural language goals~\cite{lawton2023reframer, ibarrola2023collaborative}.
Computer-based studies have shown creativity augmentation benefits of co-creation~\cite{lawton2023reframer, ibarrola2023collaborative, davis2016DrawingApprentice} since computer agents can add serendipity and reformulate user's original intentions leading to unexpected by enjoyable outputs~\cite{lawton2023toolAtool}.
However, Computer-based painting models do not transfer well out-of-the-box into the real world due to the Sim2Real gap~\cite{schaldenbrand2021contentMaskedLoss, schaldenbrand2023frida, bidgoli2020manuelPainter}. 



\subsection{Robotic Image Co-Creation}

There exists many real-world methods for robot painting and drawing~\cite{ lee2022icraSketchRobot, carter2017DarkFactoryPortraits, lindemeier2018Edavid, sola2022dreamPainter}, however,
few systems have incorporated perception into their systems to enable \canvascontinuation.
Cobbie~\cite{lin2020cobbie} is a co-drawing system that boosted ideation for novice drawers, however, it is limited to drawing on blank areas of the paper rather than engaging with the user drawn content.
\cite{shaik2021coCreativeDiffAbledKids} created a robot arm that can draw from speech inputs that are limited to simple objects found in the Quick, Draw! dataset~\cite{jongejan2016quickDraw}.
The FRIDA~\cite{schaldenbrand2023frida} system is the only system directly capable of making physical paintings that engage with existing content conditioned on natural language goals.
While FRIDA plans based on current canvas state, it uses CLIP and gradient descent for planning which produces paintings that are very noisy and only loosely resemble the input text.

\vspace{-9pt}
\section{METHOD}

Our approach, CoFRIDA shown in Fig.~\ref{fig:method}, is made up of three primary components: (1) The \Canvascontinuation Module, which produces images illustrating how the robot should add content to an existing canvas given a text description,  (2) FRIDA~\cite{schaldenbrand2023frida}, a robotic painting system for planning actions from given images, and (3) a self-supervised method for creating training data using FRIDA to fine-tune pre-trained models in the \Canvascontinuation Module.





\subsection{Self-Supervised Data Creation}\label{sec:data-creation}

\begin{figure*}[t!]
    \centering \small
    \includegraphics[width=1.\textwidth]{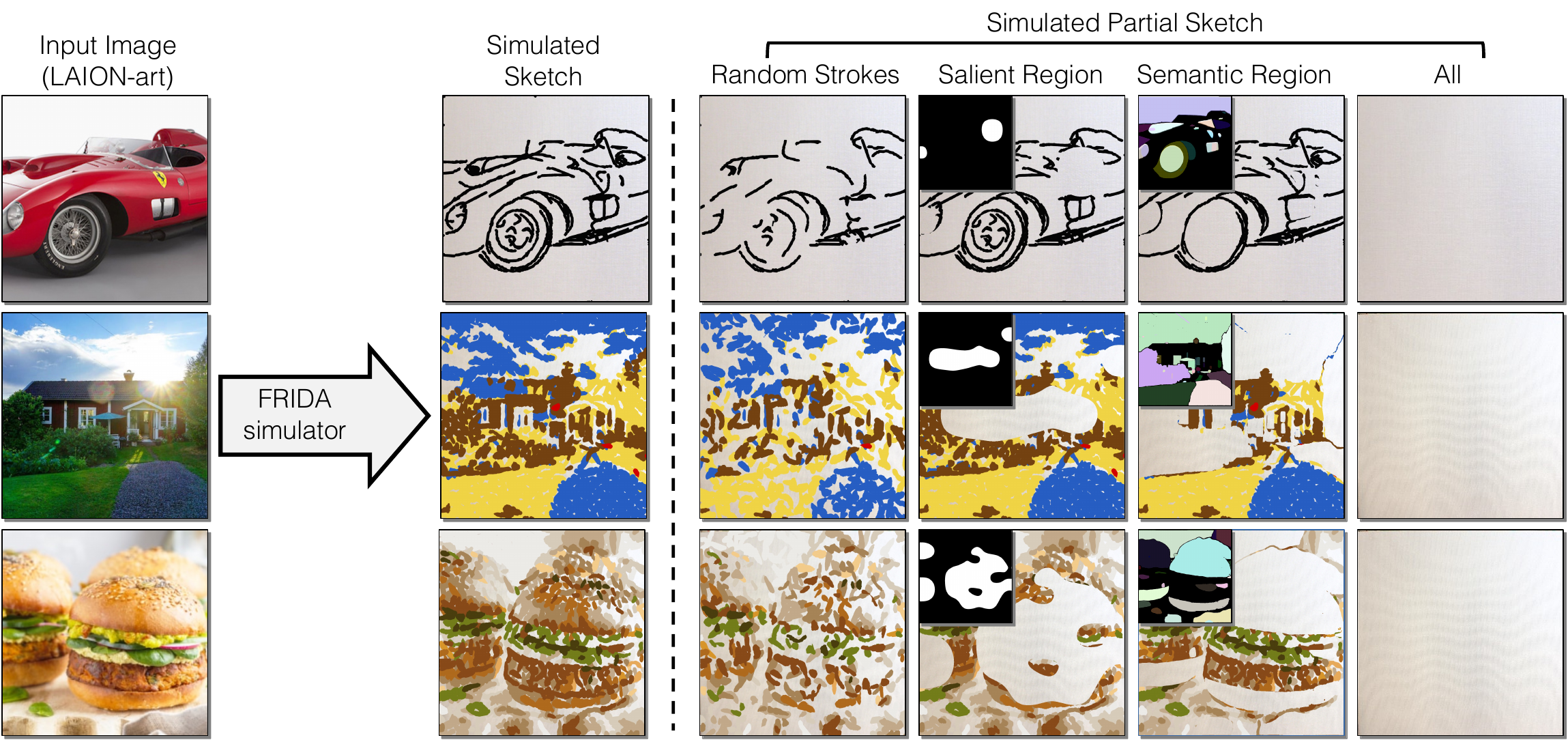}
    \vspace{-7mm}
    \caption{
    \textbf{Self-Supervised Dataset Creation.} 
    We describe the process of generating the self-supervised training data pairs for fine-tuning the \Canvascontinuation Module. We start with the input images from the LAION-art dataset and convert them into simulated sketch outputs with the FRIDA simulator. Next, we create partial sketches in four different ways: removing random strokes, removing the salient region, removing a semantic region, and removing all strokes. 
    \vspace{-6mm}
    }
    \label{fig:data_creation}
\end{figure*}

While there exist some supervised data of human-created co-paintings \cite{parikh2020exploring, codrawingsSeattle2020}, they are only on the order of tens of examples and were not made using the same materials available to our robot.
To support \canvascontinuation tasks, we propose a self-supervised method for generating training data to train a \Canvascontinuation Module. We simulate paintings of images from the art subset of the LAION image-text dataset~\cite{schuhmann2022laion5b} using FRIDA with image-guidance loss (difference of CLIP embeddings of images). To create partial paintings, strokes are removed selectively
to support a variety of \canvascontinuation tasks: remove all strokes, a random subset of strokes, strokes corresponding to a salient region (defined with CLIP as in \cite{vinker2022clipasso}) of the image, and strokes from a semantic region (using Segment Anything~\cite{kirillov2023segmentAnything}). Illustrative examples are shown in Fig.~\ref{fig:data_creation}.

Some source images cannot be accurately represented with the robot's abilities. 
We filter out such images by removing instances that have a CLIPScore between the simulated full paintings and the text less than 0.5. 

We use this self-curated data to fine-tune a base text-to-image generation model to be able to 1) continue to create content on an existing canvas and 2) generate images that the target robot is capable of painting.

\begin{figure}[t!]
    \centering \small
    \includegraphics[width=\columnwidth]{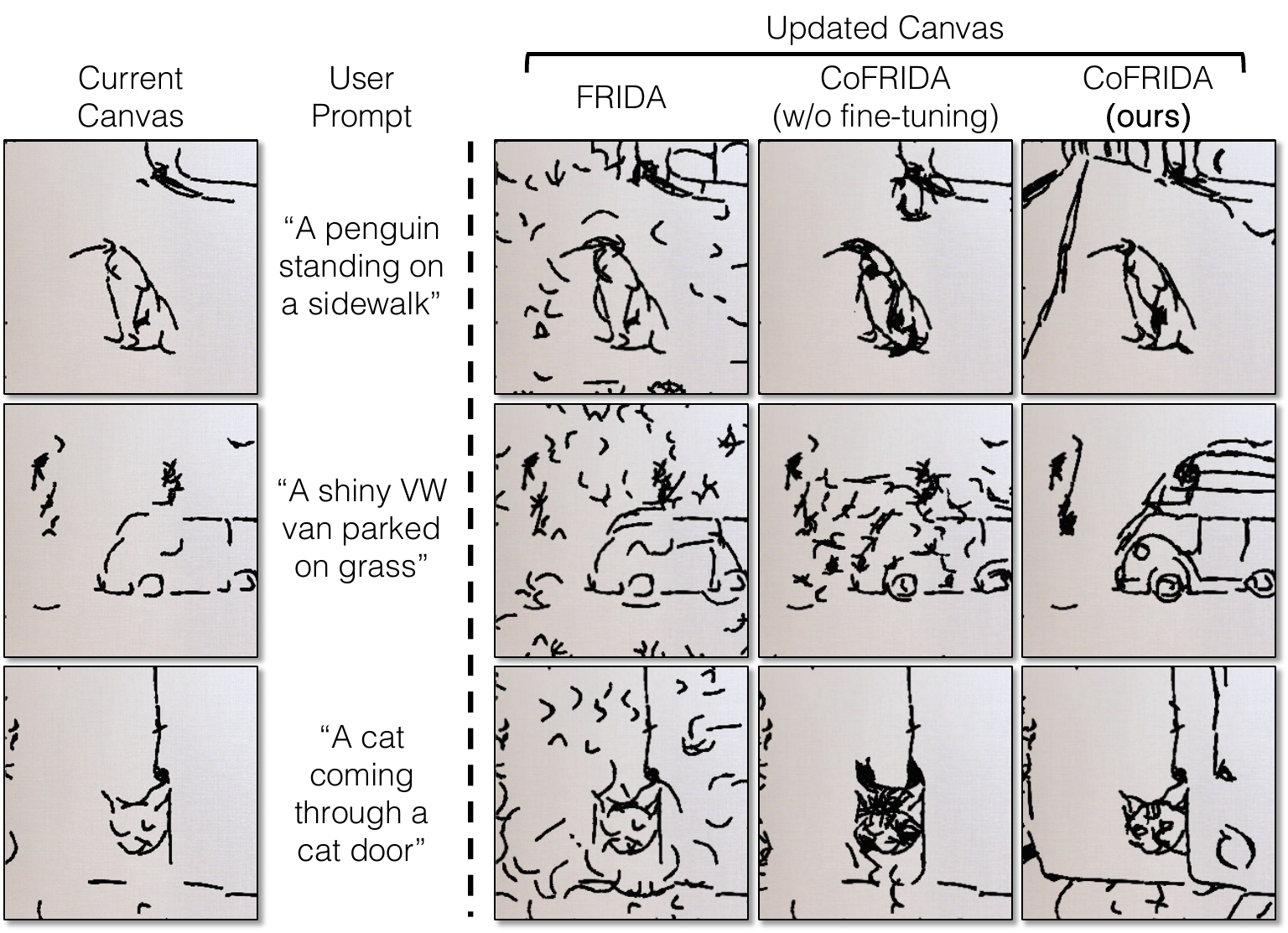}
    \vspace{-5mm}
    \caption{
    \textbf{Qualitative Comparison.} We show a comparison between three methods of performing text-based canvas updates: FRIDA, CoFRIDA without fine-tuning, and CoFRIDA with fine-tuning (ours). FRIDA uses a CLIP based optimization and generates outputs that are noisy. CoFRIDA without fine-tuning, is not aware of the constraints of the robot and  generates an output that is difficult for the robot to execute and often does not satisfy the text prompt specified by the user. In contrast, CoFRIDA outputs an updated canvas that reflects the user prompt without being noisy. 
    }
    \label{fig:study_examples}
    \vspace{-4mm}
\end{figure}

\subsection{\Canvascontinuation Module}

The goal of the \Canvascontinuation Module (Fig.\ref{fig:method}) is to generate an image of how the robot should complete the painting given a photograph of the current canvas and a user given text description.  
The \Canvascontinuation Module uses Instruct-Pix2Pix~\cite{brooks2023instructpix2pix}  as a pre-trained model as it enables conditioning the output on an input canvas. 
The pre-trained Instruct-Pix2Pix, however, has two shortcomings to be used for co-painting: (1) the generated images do not reflect actual robotic constraints, and (2) the existing canvas can sometimes be overwritten completely as shown in Fig.~\ref{fig:multiple_runs}. To overcome these limitations, 
we fine-tune Instruct-Pix2Pix using the dataset of partial and full drawings with their captions described in Sec.~\ref{sec:data-creation}. 

Fine-tuning is performed using FRIDA's simulated canvases because (1) it would be infeasible to generate a large-scale dataset with the physical robot, and (2) the \Canvascontinuation Module output is eventually used with the FRIDA simulation.

\subsection{FRIDA}

In this work, we use the FRIDA~\cite{schaldenbrand2023frida} robotic painting system for both planning actions from given images and creating self-supervision data. FRIDA uses Real2Sim2Real methodology to enable it to paint with a variety of media such as acrylic paint or markers.  Because FRIDA plans using a perceptual loss, it is capable of making, for example, drawings using markers from color photographs.

FRIDA plans actions to guide the robot to make the current canvas look like the output of the Co-Painting Module. Using FRIDA's internal simulation, we visualize a prediction of what the actions will look like, and the robot can execute the actions to update the real canvas, Fig.~\ref{fig:method}.




\section{EXPERIMENTS}

\subsection{Baselines}


We compare CoFRIDA to the original FRIDA's~\cite{schaldenbrand2023frida} CLIP-guided text-to-painting method.
We investigate the effects of our fine-tuning procedure on Instruct-Pix2Pix in the \Canvascontinuation Module by comparing our method (CoFRIDA) with pre-trained Instruct-Pix2pix (CoFRIDA w/o fine-tuning).

\subsection{Different Painting Settings \label{sec:painting_settings}}

The FRIDA painting system can paint with various brushes and can have different color constraints. We test CoFRIDA using three different painting settings (1) acrylic painting using one brush and 12 colors which can differ from painting to painting, (2) acrylic painting with a fixed 4-color palette, and (3) a black Sharpie marker. Examples of these three settings are shown in Fig.~\ref{fig:data_creation},~\ref{fig:frida_vs_cofrida}, and ~\ref{fig:semantic_sim2real_gap2} .  The robot can only be used in one of these settings at a time. However, users can paint using any media of choice, leading to mixed media paintings in Fig.~\ref{fig:out_of_distribution}.

\begin{figure}[t]
    \centering \small
    \includegraphics[width=\columnwidth]{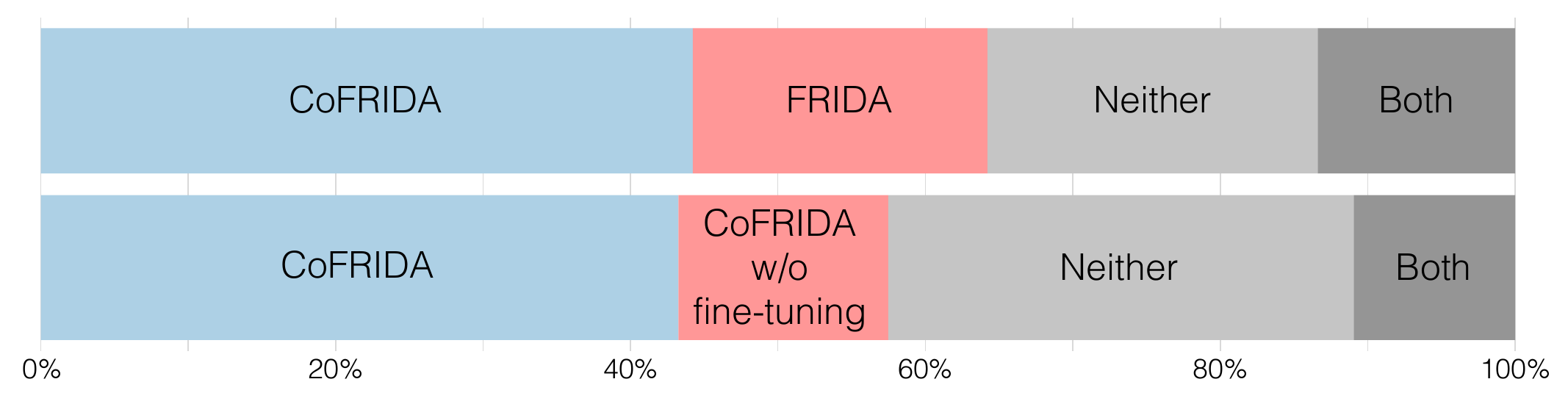}
    \vspace{-7mm}
    \caption{ 
    \textbf{User Preference Study.}
    Results from two MTurk Surveys. Presented with a text description, participants chose which of two drawings (CoFRIDA versus either FRIDA or CoFRIDA without fine-tuning) was more similar to the text, neither, or both. See Fig.~\ref{fig:study_examples} for examples.}
    \label{fig:mturk_results}
\end{figure}

\begin{table}[t!]
\centering 
\caption{CLIPScores and BLIPScores computed on robot simulated drawings (See Fig.~\ref{fig:study_examples}). Sim-to-real gap measurements, $\Delta_{pix}$ and $\Delta_{sem}$, measure the difference between the Co-Painting Module output and the simulated drawing of that image.}
\label{auto_metrics}
\begin{tabular}{@{}lcccc@{}}
\toprule
& CLIPScore $\uparrow$ & BLIPScore $\uparrow$   & $\Delta_{pix}$ $\downarrow$ & $\Delta_{sem} \downarrow$ \\ \midrule
FRIDA                                                                  & \textbf{0.741} & \textbf{0.192} & ---                                                  & ---                                                            \\ \midrule
\begin{tabular}[c]{@{}l@{}}CoFRIDA\\w/o fine-tuning\end{tabular}                                                       & 0.595          & 0.162          & 0.195                                                  & 0.241                                                            \\ \midrule 
CoFRIDA                                                                   & 0.624          & 0.178          & \textbf{0.052}                                                  & \textbf{0.035}                                                           \\ \bottomrule
\end{tabular}
\end{table}

\subsection{Evaluation}

\textbf{Text-Image Alignment} -
Two automatic methods of comparing image and text are CLIPScore~\cite{hessel2021clipscore} and BLIPScore~\cite{li2022blip}, which measure the similarity between images and text with a pre-trained image-text encoders. Because FRIDA directly optimizes the CLIPScore to create images from text, this method is unfairly advantaged when using CLIPScore. We use MTurk to achieve large-scale fair evaluation of text-image alignment.

\textbf{Semantic Sim2Real Gap} -
It is important that between the output of the \Canvascontinuation Module ($\cocreate$) and the FRIDA simulation ($\frida$) there is little loss in semantic meaning. 
A naive approach at measuring this loss is the mean-squared-error between the images' pixels (Eq.~\ref{eq:sim2real}). However, this is sensitive to low-level variation in details such as color or tone differences which are tolerable as long as the high-level content in the images is the same.
To measure the high-level difference, we propose to use the cosine distance between CLIP image embeddings, $\Delta_{sem}$, Eq.~\ref{eq:sem_sim2real}, referred to as the Semantic Sim2Real Gap.
\vspace{-3pt}
\begin{align}\vspace{-7pt}
    \Delta_{pix} &= || \cocreate - \frida ||^2_2 \label{eq:sim2real} \\
    \Delta_{sem} &= \cos(CLIP(\cocreate), CLIP(\frida)) \label{eq:sem_sim2real}
\end{align}
A proper Sim2Real gap measurement would compare the output of the \Canvascontinuation Module to the real drawing, however, it is infeasible to generate a robust number of real-world samples. Because the Sim2Real gap between the FRIDA simulation and the real drawing is the same across all tested methods, we can fairly use the FRIDA simulations in lieu of the real drawings for comparing the Sim2Real gaps of \Canvascontinuation Module variations.
\vspace{-3pt}
\section{RESULTS}

\subsection{\Canvascontinuation}

To test the ability of CoFRIDA to work with an existing canvas state, we focus on Sharpie marker drawings where no erasing is possible, forcing the model to have to adapt to and use the existing markings on the page. To create the partial drawing, we generate an image with Stable Diffusion using prompts from the PartiPrompts~\cite{yu2022partiPrompts} dataset, then simulate the drawing with just 35 strokes as depicted in Fig.~\ref{fig:study_examples}. 
We generated 40 images from different prompts per method.
CLIPScore~\cite{hessel2021clipscore}, BLIPScore, and Sim2Real gap measures are reported in Table~\ref{auto_metrics}.
Since FRIDA maximizes CLIPScore, it was expected and confirmed to have the highest CLIPScore. BLIP is also expected to correlate with CLIP, leading FRIDA to have an artificially high BLIPScore.

To properly assess the image-text similarity of the drawings from partial sketches, we conducted an MTurk survey summarized in Fig~\ref{fig:mturk_results}. 24 unique participants were shown a language description then  two images (one from ours and the other one of the two baselines, in random order). Participants were instructed to choose which image fits the given caption better, or to select neither or both. Each image pair was evaluated by 4 unique participants leading to 160 comparisons per baseline. While many participants found neither image fit the text description (an indicator of the challenging nature of co-painting), CoFRIDA was generally indicated as having clearer content over FRIDA and CoFRIDA without our fine-tuning.

In terms of the proposed Semantic Sim2Real Gap, CoFRIDA outperforms the baselines indicating that our fine-tuning guided Instruct-Pix2Pix to produce images that were less likely to change meaning when painted by FRIDA.

\begin{figure}[t!]
    \centering
    \includegraphics[width=\columnwidth]{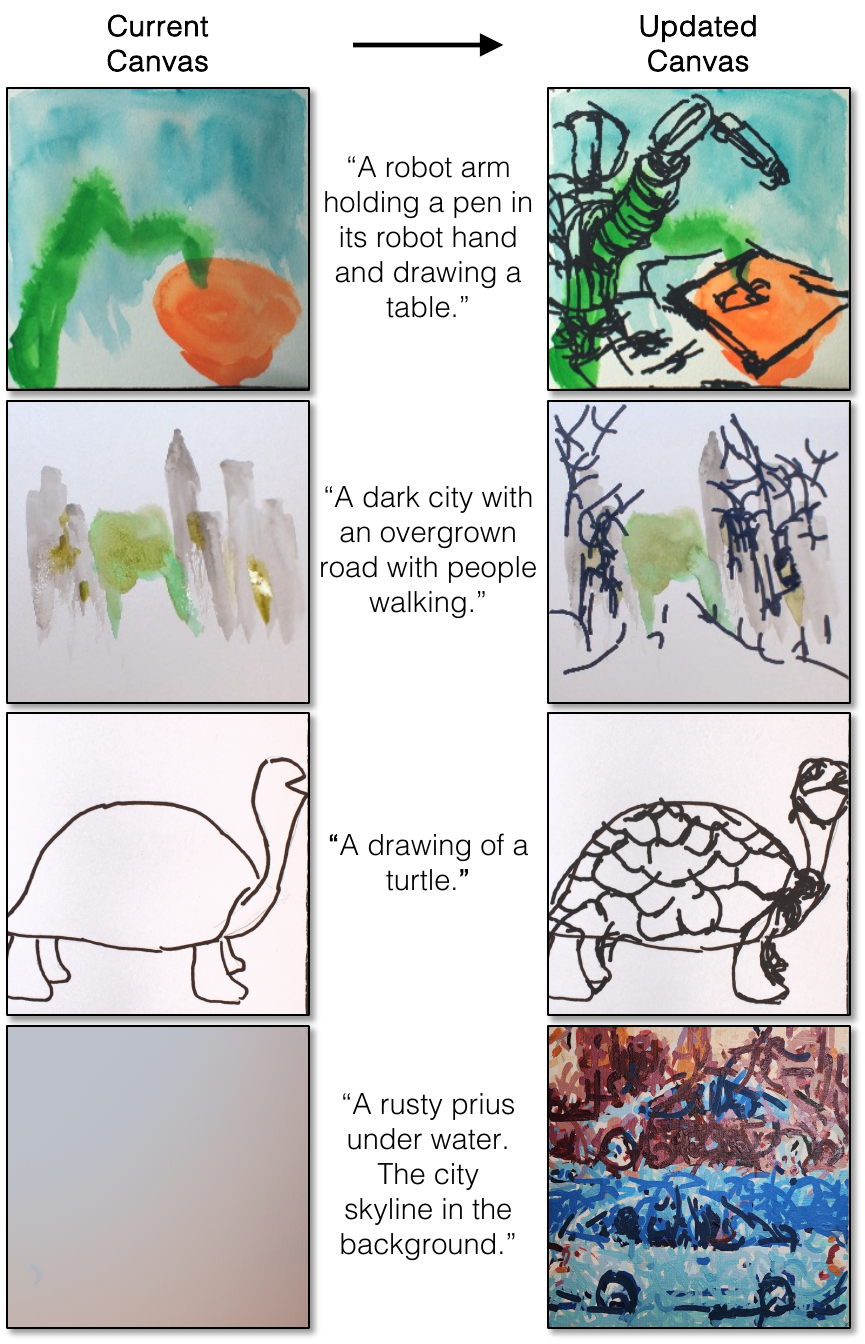}
    \vspace{-7mm}
    \caption{
    \textbf{Mixed-Media Paintings.} 
    CoFRIDA can use markers and paintbrushes to co-paint with a human.
    Despite being fine-tuned with a single medium, CoFRIDA can still perform \canvascontinuation when a user uses different media such as watercolors.}
    \label{fig:out_of_distribution}
    \vspace{-5mm}
\end{figure}

\begin{figure*}
    \centering \small
    \includegraphics[width=\textwidth]{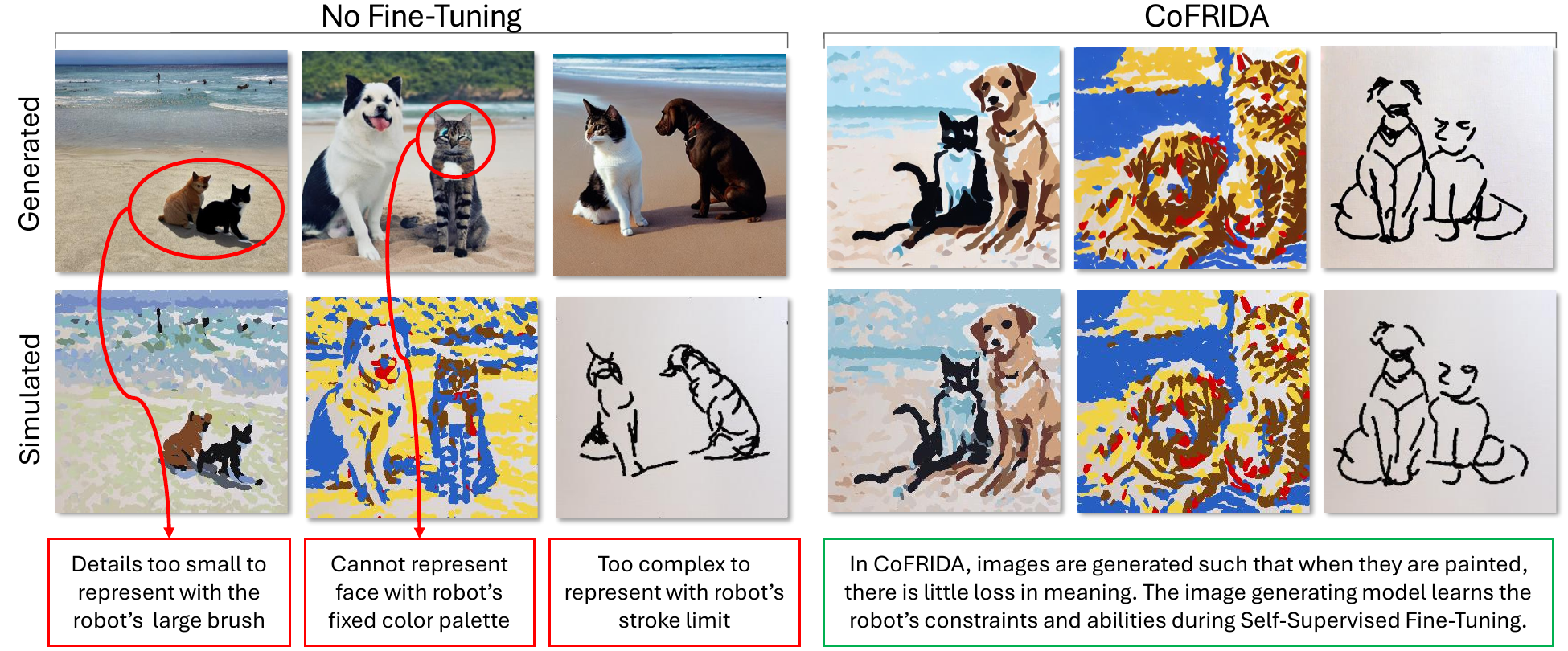}
    \vspace{-15pt}
    \caption{
    \textbf{Learning Robotic Constraints.}
    We compare images generated by a pre-trained Stable Diffusion model (left) to those generated by our proposed CoFRIDA module (right) with the prompt ``A dog and a cat sitting next to each other on the beach" in three different painting settings (Sec.\ref{sec:painting_settings}). 
    The top row shows the images generated by each of the models and the bottom row shows the corresponding FRIDA simulation. 
    }
    \label{fig:semantic_sim2real_gap2}
    \vspace{-4mm}
\end{figure*}

\subsection{Multiple Turns}

A \canvascontinuation system must be capable of accommodating multiple iterations of human-robot interaction in which the robot adds content but does not completely overwrite the human's prior work. We simulate this by having the robot create sequences of modifications to a simulated painting with different text prompts in Fig.~\ref{fig:multiple_runs}. The baseline methods tend to either avoid making changes or make huge changes to the canvas, whereas CoFRIDA makes updates that are more reasonable for the robot to achieve and integrate naturally with prior work. 

\begin{figure}
    \centering
    \includegraphics[width=\columnwidth]{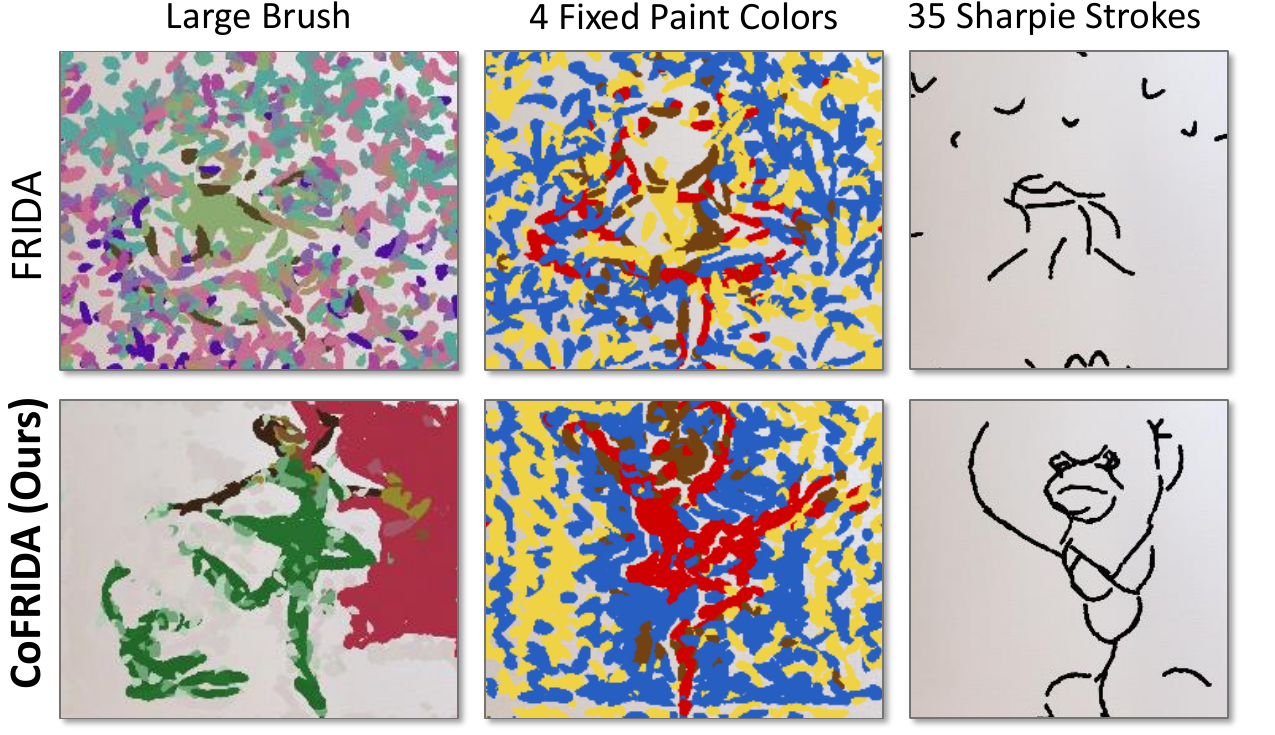}
    \vspace{-25pt}
    \caption{
    Comparing CoFRIDA's fine-tuned pre-trained image generator versus FRIDA's CLIP-guided method for generating paintings from the text ``A sad, frog ballerina doing an arabesque" in three painting settings.
    }
    \label{fig:frida_vs_cofrida}
    \vspace{-5mm}
\end{figure}

\subsection{Text Conditioned Paintings}

FRIDA's text-to-painting method relies on feedback through CLIP which results in noisy, unclear imagery. We compare CoFRIDA which uses a pre-trained generative model to FRIDA in Fig.~\ref{fig:frida_vs_cofrida}.  CoFRIDA's paintings are far more clear and capture the caption better than FRIDA in various painting settings.

\subsection{Real Paintings}

We used FRIDA's simulation to make large scale data creation and evaluation feasible. Fig.~\ref{fig:out_of_distribution} displays multiple real-world examples of CoFRIDA's drawings and paintings. CoFRIDA is able to successfully use content on canvases that is out of distribution from its fine-tuning training data as with the watercolor and marker examples in Fig.~\ref{fig:out_of_distribution}.









\vspace{-5pt}
\section{DISCUSSION}

\textbf{Limitations and Ethical Considerations} 
CoFRIDA stands out as a successful collaborative painting system, but is limited to discrete turn-taking interactions. While our self-supervised training data creation method (Fig.~\ref{fig:data_creation}) was informed by real co-painting data, a more end-to-end approach where the system learns how to form the partial paintings could result in even better results.

In light of recent discoveries of harmful content in the LAION dataset, we have inspected the subset of data used to fine-tune the models used in this paper and have not found harmful content. We have changed the code to use the COCO dataset~\cite{lin2014coco} and have not seen a degradation in the quality of results.
CoFRIDA is subject to the biases of Stable Diffusion~\cite{rombach2021stableDiffusion} and its training data~\cite{birhane2021laionProblems}, and so we recommend the usage of CoFRIDA with caution and solely for research purposes.  

\textbf{Learning Robotic Abilities}
Our self-supervised fine-tuning procedure guided the pre-trained model to generate images that, at a pixel-level, appeared similar to what FRIDA can paint, but is it learning the actual robot constraints or just a low-level style transfer? We computed the Sim2Real gap measurements between the LAION images and their FRIDA simulations (as seen in Fig.~\ref{fig:data_creation}) along with the CLIPScore of the simulation and text prompt. We found that $\Delta_{pix}$ had a small, insignificant Pearson correlation ($-0.08$, $0.08$ $p$-value) with the CLIPScore of the painting whereas $\Delta_{sem}$ had a significant, negative correlation ($-0.48$, $2.4e-31$ $p$-value). Because CoFRIDA greatly decreases the $\Delta{sem}$, this indicates that CoFRIDA's fine-tuning technique is not solely changing the low-level appearance (akin to style-transfer) over the output of its base model. It appears that CoFRIDA is learning the robot's abilities, as seen in Fig.~\ref{fig:semantic_sim2real_gap2} where CoFRIDA's Co-Painting Module produces images with (1) very prominent and clear content, when the robot's brush is large (2) select and limited colors, when the robot paints with fixed palettes or markers, and (3) sparse, concise drawings when the number of strokes is limited.

\section{CONCLUSIONS}

An end-to-end approach, like FRIDA, that optimizes the brush strokes towards the text goal tends to produce noisy looking paintings that only loosely resemble the text because it operates in a low-level space without a global context. Additionally, it is hard to incorporate interactivity beyond an initial input.
We present Collaborative FRIDA (CoFRIDA), a hierarchical approach for interactive human-robot co-painting where semantic planning via pre-trained models happens in a high-level, pixel space before being transferred to a low-level brush stroke planner. 
Pre-trained models do not immediately provide the requirements for co-painting, as they do not know the capabilities of the robot. Whereas the Real2Sim2Real methodology improves low-level action-space planning in FRIDA, 
the proposed self-supervised fine-tuning procedure provides a method for adapting powerful pre-trained models for high-level robotic planning. CoFRIDA uses this hierarchical approach for reducing the Sim2Real gap, achieving enhanced performance over the baselines.

\section{Acknowledgments}
This work was partly supported by NSF IIS-2112633, the Packard Fellowship, and the Technology Innovation Program (20018295, Meta-human: a virtual cooperation platform for a specialized industrial services) funded By the Ministry of Trade, Industry \& Energy (MOTIE, Korea).

\bibliographystyle{IEEEtran}
\bibliography{IEEEabrv,ref}

\end{document}